\documentclass[journal]{IEEEtran}
\ifCLASSINFOpdf
\else
\fi
\usepackage{amsmath}
\usepackage{algorithmic}

\usepackage{graphicx}
\usepackage{subfigure}
\usepackage{url}

\usepackage{amsmath}

\usepackage{algorithmic}

\usepackage{array}

\usepackage{graphicx}
\usepackage{subfigure}
\graphicspath{ {./} }

\usepackage{booktabs} 
\usepackage{multirow} 
\usepackage{amssymb}
\usepackage{bbding}

\begin{document}
%
\title{MobiAct: Efficient MAV Action Recognition Using MobileNetV4 with Contrastive Learning and Knowledge Distillation}
%
%
%

 \author{Nengbo Zhang$^{1}$, 
         Hann~Woei~Ho$^{1,*}$,~\IEEEmembership{Senior Member,~IEEE,}
 \thanks{$^{1}$Zhang Nengbo, Hann Woei Ho are with School of Aerospace Engineering, Engineering Campus, Universiti Sains Malaysia, 14300 Nibong Tebal, Pulau Pinang, Malaysia (email: zhangnb@student.usm.my; aehannwoei@usm.my).}

 \thanks{*Corresponding author: Hann~Woei~Ho (email: aehannwoei@usm.my) }
 }
%
%

\markboth{IEEE Transactions on Artificial Intelligence, Submitted}%
{Shell \MakeLowercase{\textit{et al.}}: Bare Demo of IEEEtran.cls for IEEE Journals}
%



\maketitle

\begin{abstract}
    Accurate and efficient recognition of Micro Air Vehicle (MAV) motion is essential for enabling real-time perception and coordination in autonomous aerial swarm. However, most existing approaches rely on large, computationally intensive models that are unsuitable for resource-limited MAV platforms, which results in a trade-off between recognition accuracy and inference speed. To address these challenges, this paper proposes a lightweight MAV action recognition framework, MobiAct, designed to achieve high accuracy with low computational cost. Specifically, MobiAct adopts MobileNetV4 as the backbone network and introduces a Stage-wise Orthogonal Knowledge Distillation (SOKD) strategy to effectively transfer MAV motion features from a teacher network (ResNet18) to a student network, thereby enhancing knowledge transfer efficiency. Furthermore, a parameter-free attention mechanism is integrated into the architecture to improve recognition accuracy without increasing model complexity. In addition, a hybrid loss training strategy is developed to combine multiple loss objectives, which ensures stable and robust optimization during training. Experimental results demonstrate that the proposed MobiAct achieves low-energy and low-computation MAV action recognition, while maintaining the fastest action decoding speed among compared methods. Across all three self-collected datasets, MobiAct achieves an average recognition accuracy of 92.12\%, while consuming only 136.16 pJ of energy and processing recognition at a rate of 8.84 actions per second. Notably, MobiAct decodes actions up to 2 times faster than the leading method, with highly comparable recognition accuracy, highlighting its superior efficiency in MAV action recognition.
\end{abstract}


%
\IEEEpeerreviewmaketitle

\section{Introduction}
Micro Air Vehicles (MAVs) play an increasingly important role in areas, such as surveillance, disaster response, and collaborative swarm operations \cite{lei2023bio}. With the increasing proliferation of multi-agent MAV systems \cite{zhang2025multirate}, reliable and secure communication among MAVs has become a critical requirement. Although traditional wireless communication remains widely used, it suffers from susceptibility to interference, bandwidth limitations, and potential security vulnerabilities. As demonstrated in recent work \cite{li2024behavior}, MAV motion behaviors can also act as a form of communication signal, enabling intent recognition among MAV agents. To overcome these issues, vision-based communication, in which MAVs exchange information through interpretable motion patterns, has emerged as a promising alternative. Compared to conventional radio-based methods, vision-based approaches provide higher resistance to external interference and enhanced transmission security. However, the realization of such communication fundamentally depends on accurate recognition of MAV motion patterns, which serve as the symbolic basis for information transmission. 
In this paradigm, each MAV action is abstracted as a motion primitive that represents a specific symbol or command. These primitives are sequentially combined to form communication codes, which can be further organized into higher-level command sequences to deliver messages among MAVs. Therefore, reliable MAV action recognition serves as the core enabling technique for efficient visual communication. Despite its significance, research on MAV action recognition remains limited.



\begin{figure}
\centering 
\includegraphics[width=0.49\textwidth]{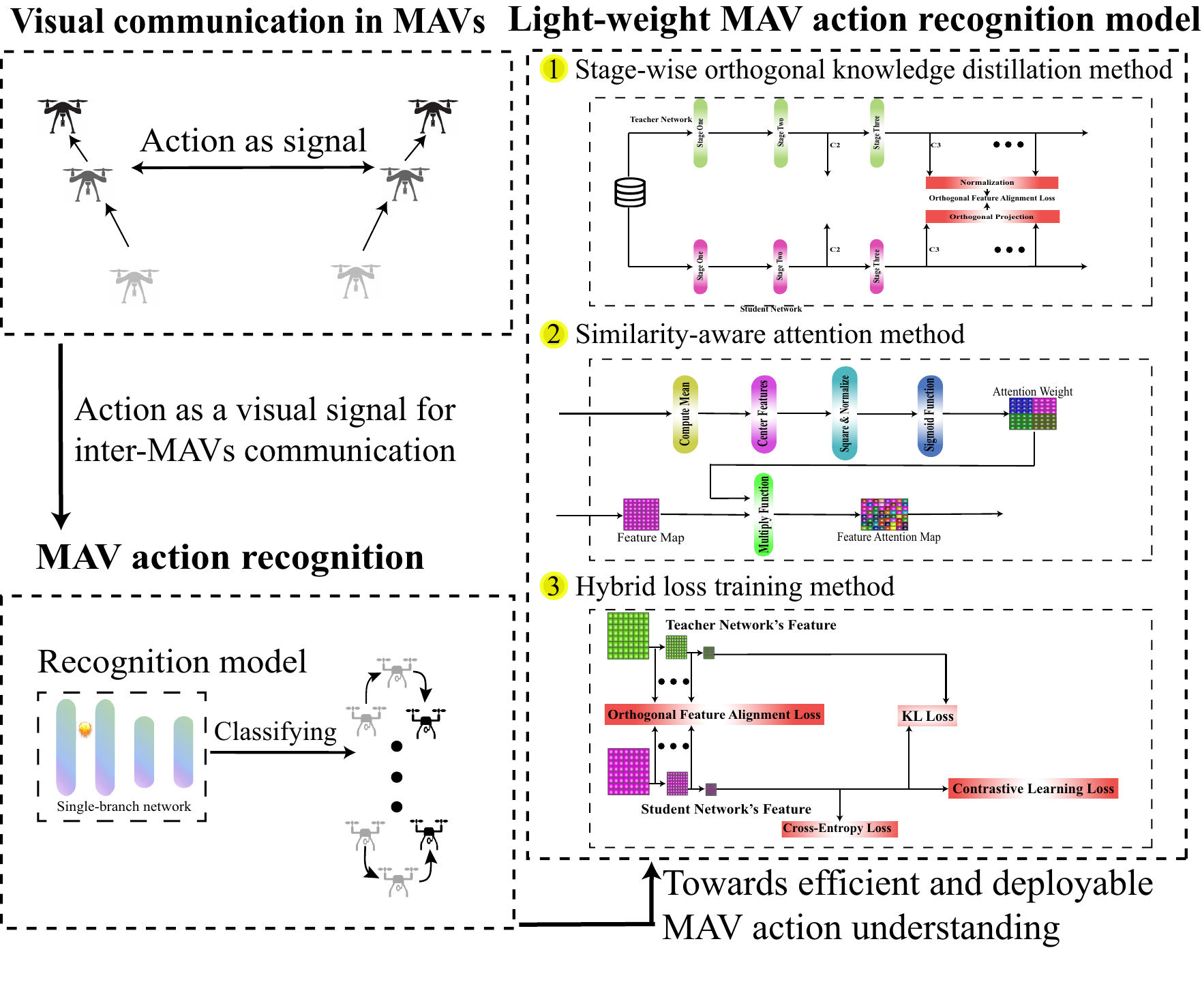}
\caption{Schematic overview of our proposed model for visual communication: Actions serve as visual signals for inter-MAVs coordination, processed via a recognition model employing (1) stage-wise orthogonal knowledge distillation, (2) similarity-aware attention module, and (3) hybrid training strategy, enabling efficient and deployable MAV action understanding.} \label{fig:MAV_comTech}
\end{figure}

In recent years, tremendous progress has been achieved in video-based human action recognition, driven by deep learning advancements. Early studies employing 3D Convolutional Neural Networks (3D CNNs) \cite{tran2015learning} effectively captured spatiotemporal dynamics from video sequences. Subsequent works further enhanced recognition accuracy by modeling spatial \cite{xu2017r} and temporal \cite{liu2018t} dependencies, while attention- and transformer-based architectures \cite{bertasius2021space,xing2023svformer} pushed performance to new levels. However, these approaches are typically designed for human-scale actions and rely on high-capacity architectures that are impractical for embedded or energy-limited systems. In contrast, MAV action recognition introduces unique challenges due to the small physical size of MAVs, their rapid and complex motion trajectories, and the diverse environmental conditions under which they operate. Although a few recent works explored multi-view learning \cite{zhang2025mavrnetrobustmultiviewlearning} and event-based sensing \cite{nengbo2025mocommotionbasedintermavvisual} for MAVs, the field is still in its infancy, and existing solutions struggle to achieve both high recognition accuracy and deployment efficiency.

Despite the remarkable advances in human action recognition field, directly transferring these techniques to MAVs remains non-trivial. Two intertwined challenges fundamentally limit the performance of existing models. First, robust recognition across observation scales is difficult to achieve. In practical settings, MAVs are recorded at varying distances, resulting in drastic changes in their apparent scale, motion amplitude, and background complexity. When the observation distance increases, motion cues become weaker and are easily disrupted by dynamic backgrounds, occlusions, or blur. These factors greatly challenge the robustness and generalization of MAV action recognition. Consequently, conventional models often fail to maintain stable recognition accuracy across scales. Second, lightweight and efficient recognition for visual communication is essential. MAV-based visual communication requires models that can decode motion patterns efficiently while operating under strict power, memory, and computational constraints. Heavy architectures and multi-modal fusion methods hinder real-world deployment and reduce communication transmission efficiency. This is because slow recognition speed causes delayed MAV responses, which in turn significantly degrades the efficiency of visual communication. Addressing these two challenges, i.e., robust recognition under scale variation and efficient decoding under limited resources, is crucial for enabling practical MAV visual communication systems.

\begin{figure*}
		\centering 
		\includegraphics[width=0.95\textwidth]{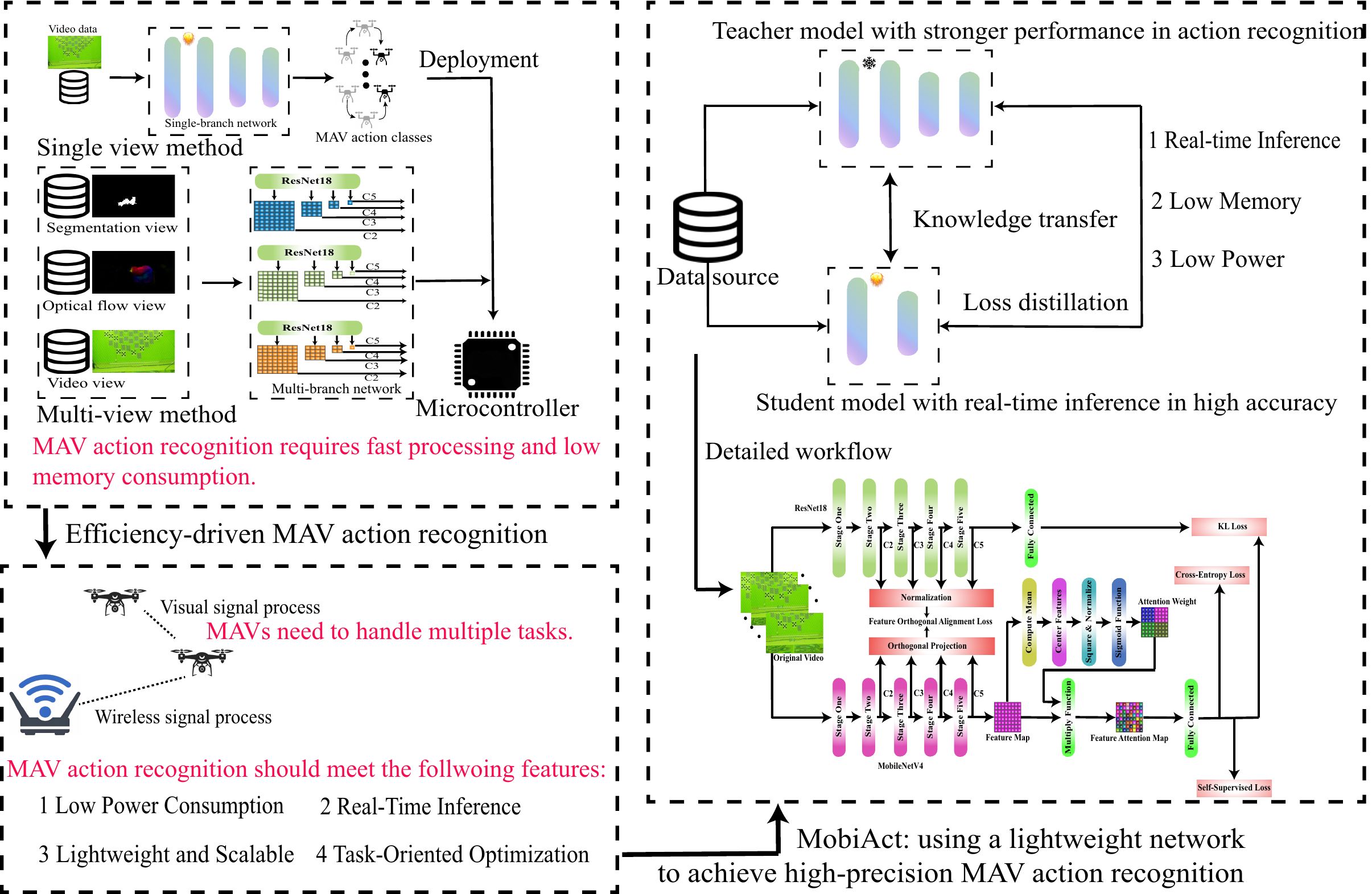}
		\caption{Schematic diagram of the MobiAct architecture. Single-view or multi-view MAV action recognition methods, when executed on microprocessors, face challenges, such as latency and memory bottlenecks. To address the efficiency challenges in MAV action recognition, we leverage lightweight neural networks for high-performance deployment.  } 
        \label{fig:firstImage}
\end{figure*}

To address these challenges, we present MobiAct, a lightweight MAV action recognition framework (illustrated in Fig. \ref{fig:MAV_comTech} and Fig. \ref{fig:firstImage}) that achieves an optimal balance among accuracy, latency, and robustness, enabling efficient and reliable visual communication. To the best of our knowledge, this is the first work to explicitly investigate the implementation of efficient MAV action recognition within a lightweight neural network paradigm. Specifically, we adopt MobileNetV4 \cite{qin2024mobilenetv4} as the backbone of the student network, enabling efficient feature extraction with reduced computational cost. To further enhance the discriminative capability of the lightweight model, we introduce a knowledge distillation strategy, where a ResNet18 \cite{he2016deep} teacher network guides the student network (MobileNetV4) to learn informative spatiotemporal representations from MAV motion sequences. In addition, we employ supervised contrastive learning to train the student network, thereby improving its ability to generalize across different MAV action categories and reducing overfitting to small-scale training data. To refine feature selection, a parameter-free attention mechanism (Similarity-aware attention module) is incorporated, which allows the model to focus on key motion cues while maintaining training stability. Finally, extensive experiments on three laboratory-collected datasets validate the effectiveness of the proposed approach, showing that it achieves accurate recognition of MAV actions while maintaining high computational efficiency and small power consumption. Therefore, the key contributions of this work are summarized as follows:

\begin{enumerate}
\item A compact neural network, MobiAct, is developed based on MobileNetV4 backbone to achieve efficient MAV action recognition with minimal accuracy degradation.
\item A stage-wise orthogonal knowledge distillation (SOKD) method is proposed. Unlike conventional distillation techniques, SOKD performs orthogonal projection of student features to align with the normalized teacher representations across multiple stages, facilitating progressive knowledge transfer.
\item A parameter-free attention mechanism (similarity-aware attention module) is integrated into the lightweight architecture to enhance the network’s ability to focus on small MAV motion cues, while a hybrid loss training strategy ensures stable optimization. Furthermore, extensive experiments on three MAV action recognition datasets of different scales are conducted to systematically validate the effectiveness of the proposed framework.
\end{enumerate}

In the subsequent sections, Section {\uppercase\expandafter{\romannumeral2}} introduces related works, including visual action recognition techniques, MAV communication development, and knowledge distillation techniques in video action recognition. Section {\uppercase\expandafter{\romannumeral3}} provides a detailed explanation of the proposed model and methods. Section {\uppercase\expandafter{\romannumeral4}} describes our experimental setups and several key results. Finally, Section {\uppercase\expandafter{\romannumeral5}} draws a conclusion and summarizes the key points of this paper.

\section{Related works}
In this section, the recent development of knowledge distillation techniques in video action recognition is first presented. Then, the overview of MAV communication development is described. Lastly, the representative works on video action recognition in recent years are discussed. 

\subsection{Knowledge Distillation for Video Action Recognition}
Knowledge Distillation (KD), an effective model compression technique, transfers knowledge from complex neural networks to lightweight models, addressing the challenge of achieving low latency and high recognition performance in resource-constrained environments, such as mobile or aerial devices. It is widely applied in computer vision tasks and terminal devices \cite{lin2022device}. A study \cite{wang2024generative} introduced a generative model-based feature KD framework for video action recognition, using attention mechanisms to transfer temporal-spatial semantics, enabling efficient compression of 3D-CNNs. Another work \cite{li2025sample} proposed sample-level adaptive KD, dynamically adjusting distillation based on sample difficulty to enhance compression in action recognition models. In sensor-based human activity recognition \cite{xu2023contrastive}, a contrastive KD approach with regularized knowledge was developed to compress deep models while preserving performance in dynamic scenarios. A multi-teacher KD framework \cite{wu2019multi} for compressed video action recognition aggregates knowledge from multiple sources to reduce model size for video tasks. Further, a part-wise KD method \cite{liu2024enhancing} for low-quality skeleton data in action recognition transfers discriminative features across body parts, supporting compression in noisy environments. Extending to heterogeneous setups, mutual KD \cite{xiao2025heterogeneous} between models was explored for wearable activity recognition, facilitating lightweight deployment. A review \cite{habib2024comprehensive} highlighted KD applications in video tasks, including action recognition with vision transformers, for model simplification in resource-limited settings. These advancements in feature-level and multi-layer KD, including orthogonal projections to minimize redundancy, pave the way for our lightweight MAV action recognition network, which employs multi-layer orthogonal feature KD to compress spatio-temporal models while retaining essential motion cues for efficient on-device processing in micro-aerial vehicles.

\subsection{The Development of MAV Communication Techniques}

Flight vehicle communication techniques play a key role in MAV swarm control, enabling coordination and data exchange. Early approaches leveraged visual sensors for formation control and data association, such as using multiple aerial cameras \cite{aranda2015formation} to guide mobile robots into desired formations while addressing limited communications and mutual perception challenges. For instance, distributed strategies \cite{montijano2013distributed} for data association in robotic networks with cameras and constrained bandwidth have been explored to ensure consistent feature matching across agents. Subsequent advancements \cite{talamali2021less} focused on optimizing communication constraints to enhance adaptability, demonstrating that restricted radio frequency interactions in local regions can improve swarm responsiveness to environmental changes, as seen in robot swarms adapting better with limited connectivity. Biologically inspired methods \cite{dong2023social}, drawing from social signaling in honey bees, have further innovated communication paradigms by emphasizing efficient information transfer in dynamic settings. To address coverage limitations, distributed learning approaches \cite{gao2022coverage} have been proposed for MAV swarm networks, mitigating signal attenuation and noise to achieve high-coverage control through adaptive positioning and routing. Networked multi-agent systems \cite{calvo2024networked} for on-demand wireless infrastructure exemplify this by dynamically deploying relays to extend swarm reach and optimize data flows. In terms of communication modes, reconfigurable antennas enable point-to-point links \cite{liang2019reconfigurable} with switchable beams tailored for aerial drone applications, while surveys of air-to-ground propagation modeling \cite{khawaja2019survey} highlight challenges in ground-to-air channels. GPS-denied navigation techniques \cite{hashim2021gps}, incorporating nonlinear stochastic observers for attitude, position, and velocity estimation, expand operational flexibility in denied environments. Efficiency gains have been realized through novel MIMO-based codebooks \cite{kim2024codebook} designed for air-to-air systems in urban air mobility corridors, significantly boosting throughput in dense scenarios. These advancements underscore the evolution toward more robust and versatile MAV communication, paving the way for innovative visual communication paradigms. Visual communication leverages onboard cameras to transmit information via observable actions or gestures, bypassing traditional radio dependencies and enabling silent, low-power operations in contested environments. At its core, MAV action recognition serves as a critical enabler for this modality, allowing swarms to interpret and respond to visual cues in real time. Therefore, future applications of lightweight MAV action recognition models hold substantial potential in realizing seamless visual communication.

\begin{figure}
\centering 
\includegraphics[width=0.5\textwidth]{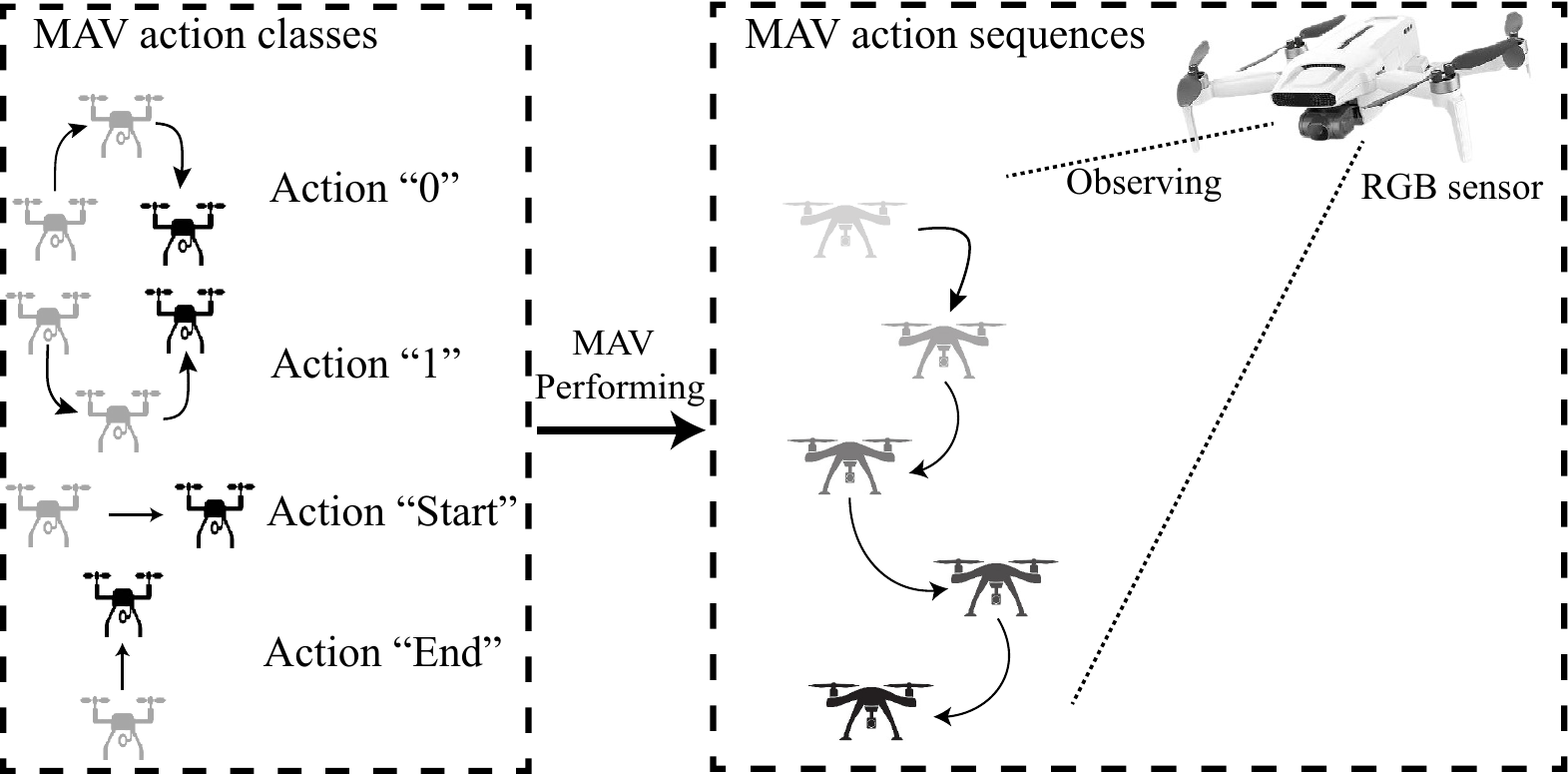}
\caption{MAV action category definition, including actions ``start'', ``end'', ``1'', ``0''.} \label{fig:MAVfigure}
\end{figure}

\subsection{Video Action Recognition Techniques}
Within the field of computer vision, video action recognition plays a pivotal role, enabling key applications in human activity analysis and micro aerial vehicle (MAV) action interpretation. In human action recognition, 3D Convolutional Neural Networks (3D CNNs) \cite{tran2015learning} have been instrumental in extracting spatio-temporal features from video sequences, building on the success of 2D CNNs in image classification. These networks jointly model spatial and temporal information, enabling end-to-end action recognition pipelines \cite{huang2023review}. However, 3D CNNs face challenges due to high computational costs and large parameter counts. To address this, R-C3D \cite{xu2017r} employs shared convolutional features between proposal and classification pipelines, generating candidate temporal action regions for efficient classification. Similarly, T-C3D \cite{liu2018t} integrates residual 3D CNNs with hierarchical temporal encoding to capture multi-granularity features, reducing computational complexity. Additionally, parameter-efficient methods, such as collaborative spatio-temporal encoding with 2D CNNs and shared layers, have been proposed to extract complementary video features \cite{li2019collaborative}. Recent advancements also include transformer-based models, such as those in space-time attention \cite{bertasius2021space} and SVFormer \cite{xing2023svformer}, which leverage attention mechanisms to capture long-range temporal dependencies and contextual relationships in human action sequences. In the domain of MAV action recognition, unique challenges arise due to subtle motion cues influenced by ego-motion, scale variation, and inter-agent synchronization. Recent studies have made significant strides in addressing these issues. For instance, event cameras have been utilized to achieve robust action recognition for MAVs \cite{nengbo2025mocommotionbasedintermavvisual}. Additionally, multi-view learning approaches \cite{zhang2025mavrnetrobustmultiviewlearning} have been developed to extract rich view information from video data, enabling high-precision MAV action recognition. These advancements highlight the evolving landscape of video action recognition, tailored to both human and MAV contexts.

\section{The proposed method}
In this section, we provide a detailed explanation of each component of our model. Section A describes the whole pipeline of the proposed model. Section B introduces stage-wise orthogonal knowledge distillation strategy and Section C provides a type of parameter-free attention mechanism. Finally, Section D shows hybrid loss in final model training.


\begin{figure*}
		\centering 
		\includegraphics[width=0.95\textwidth]{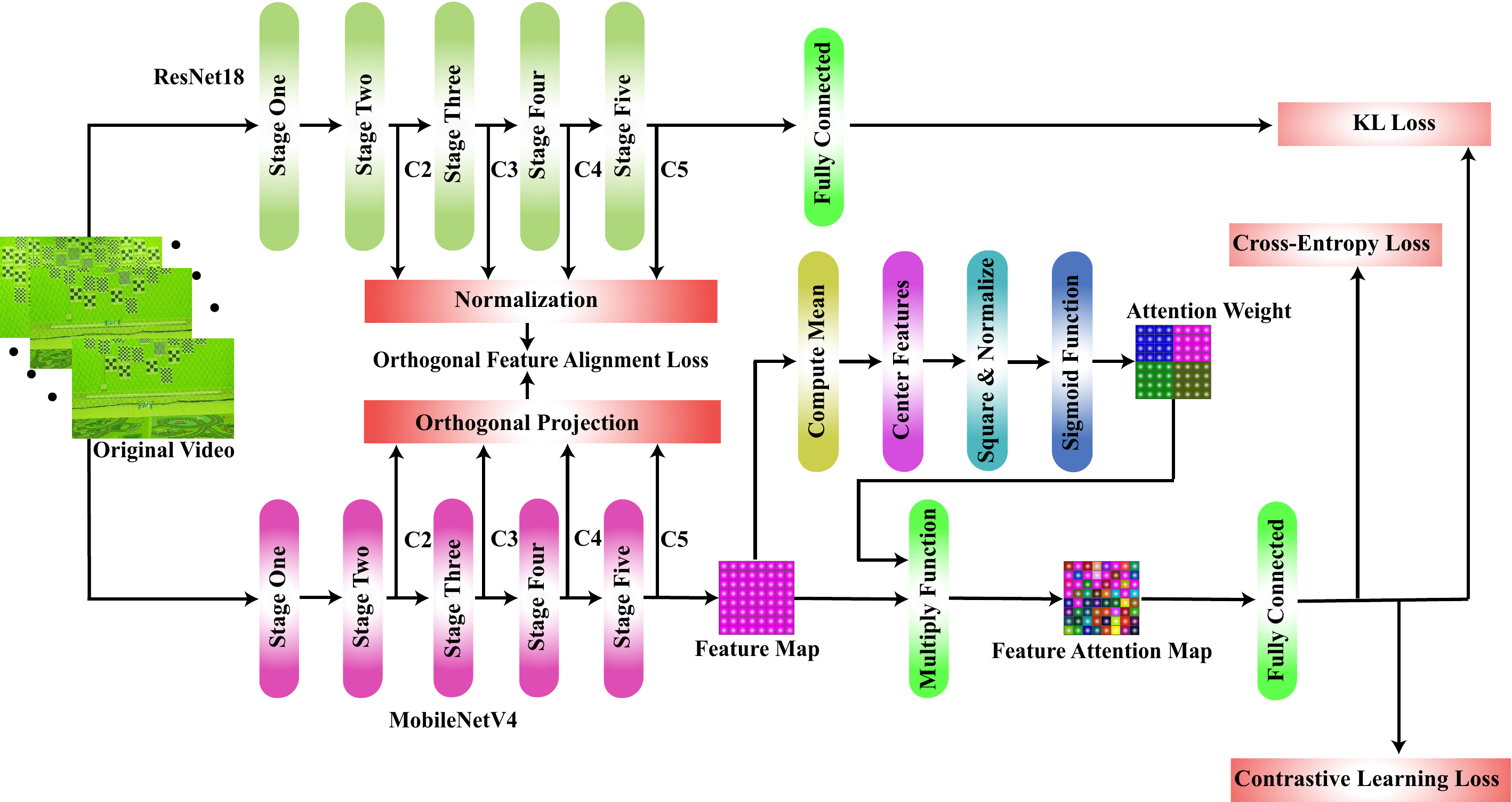}
            
		\caption{Overview of the proposed MobiAct framework. The MAV action video is processed in parallel by the teacher network (ResNet18) and the student network (MobileNetV4). Their high-level features are aligned through orthogonal projection and optimized using the Orthogonal Feature Alignment module. In parallel, a Similarity-Aware Attention (SAA) mechanism enhances discriminative feature representation. The student network is trained under a hybrid learning strategy guided by the teacher model. } \label{fig:pipe}
\end{figure*}

\subsection{Overall architecture of the proposed MAV action recognition model}


To achieve efficient and robust MAV action recognition for four motion patterns (illustrated in Fig. \ref{fig:MAVfigure}), we design a lightweight teacher–student framework that balances recognition accuracy and computational efficiency. As illustrated in Fig. \ref{fig:pipe}, the framework integrates three main components: Stage-wise Orthogonal Knowledge Distillation (SOKD) for progressive feature transfer, a Similarity-Aware Attention (SAA) module for lightweight attention, and a hybrid training objective that combines contrastive learning with supervised losses. Together, these components enable the framework to achieve competitive performance while maintaining low computational overhead, making it well suited for real-time MAV swarm deployment.

The backbone of the framework adopts ResNet-18 as the teacher network, which extracts rich spatio-temporal features from multiple stages (C2–C5) of MAV videos. These features capture subtle motion dynamics, such as hovering, dodging from MAV motion video. The student network is a lightweight MobileNetV4 \cite{qin2024mobilenetv4}, designed for edge devices with limited resources. To reduce the performance gap between teacher and student, intermediate features are aligned using orthogonal projection, allowing the student to inherit discriminative capabilities without incurring the heavy computational cost of the teacher.

At the core of the framework is the SOKD module, which performs stage-wise alignment between corresponding features of the teacher and student networks. This progressive alignment enforces consistency across multiple network depths, ensuring the student learns robust representations that generalize well to challenging aerial scenarios, including occlusions, scale variations, and motion blur. By employing orthogonal projection, the distillation process reduces redundancy and enhances the transfer of complementary MAV motion information.

To further enhance the quality of student features, the framework incorporates SAA module, a parameter-free attention mechanism that leverages the local self-similarity of spatiotemporal feature maps. Unlike conventional attention modules that rely on additional learnable parameters, SAA module adaptively highlights motion-relevant regions, such as propeller dynamics and trajectory variations, while suppressing irrelevant background noise. This design not only improves recognition accuracy but also preserves the lightweight nature of the student network, making it particularly suitable for real-time MAV action recognition where efficiency and low computational cost are critical.

The training process is guided by a hybrid objective that integrates contrastive learning \cite{gui2024survey}, cross-entropy loss, orthogonal feature alignment loss, and KL-divergence. Contrastive learning promotes consistency in feature representations across different views of the same action, thereby improving robustness against viewpoint shifts, lighting changes, and other variations common in MAV action recordings. Cross-entropy loss ensures accurate action classification, while KL-divergence aligns the student’s prediction distribution with the teacher’s soft labels, improving the effectiveness of knowledge transfer. This combination of objectives allows the framework to leverage both discriminative supervision and structural feature alignment, leading to stronger performance under limited resources.

In summary, the proposed framework combines progressive feature distillation,  parameter-free attention, and a hybrid training strategy to achieve efficient and accurate MAV action recognition. It reduces dependence on large labeled datasets, improves generalization under challenging aerial conditions, and delivers low-latency performance, making it particularly suitable for deployment in real-time MAV swarm scenarios.

\subsection{Stage-wise Orthogonal Knowledge Distillation on MAV action classification }
\begin{figure}
\centering 
\includegraphics[width=0.45\textwidth]{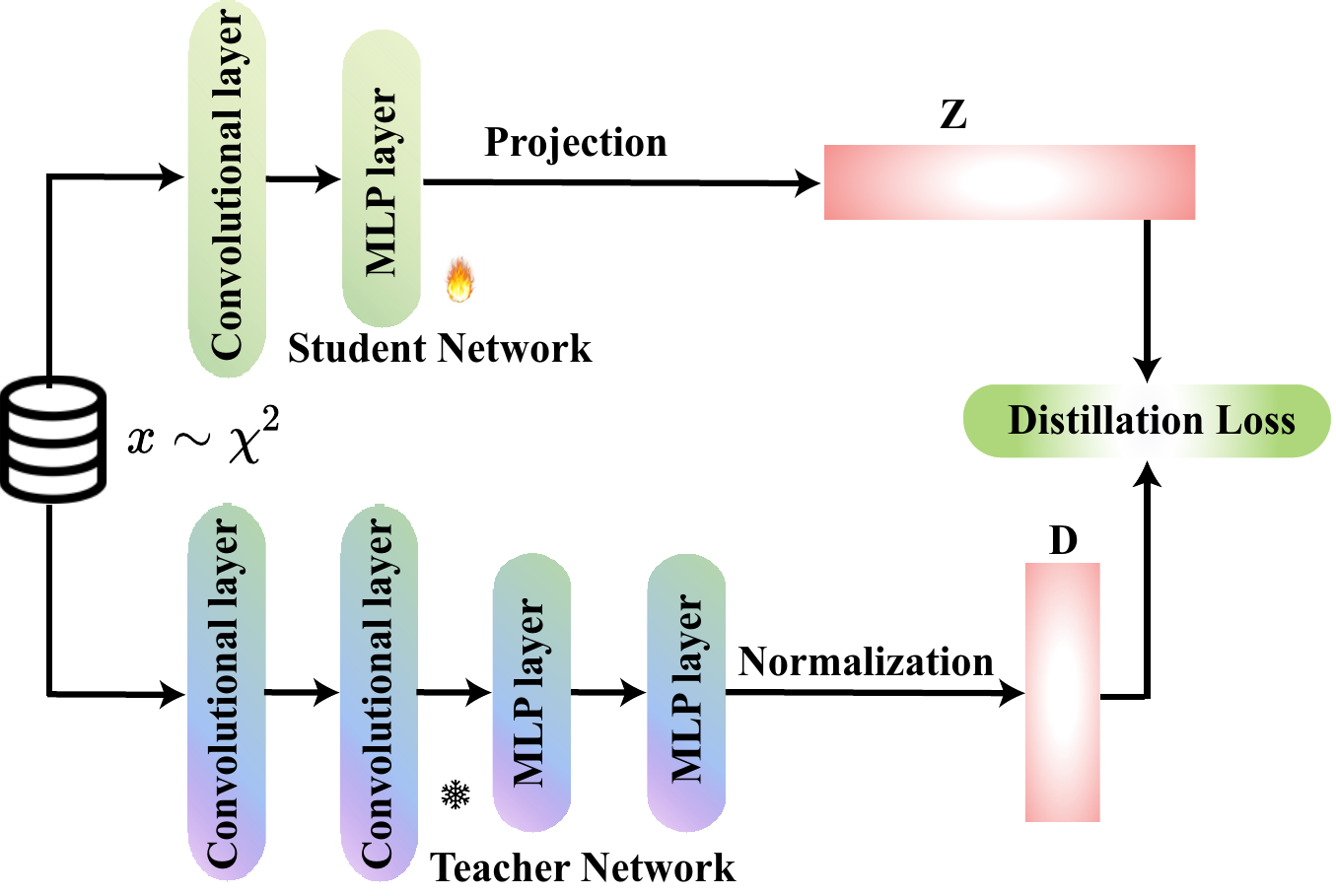}
\caption{The overview of basic knowledge distillation. The student network generates projected features $Z$, while the teacher network produces normalized features $D$. 
The distillation loss is computed between $Z$ and $D$ to transfer knowledge from the teacher to the student.} \label{fig:KDfigure}
\end{figure}

In order to achieve transferring robust knowledge of MAV actions, we introduce orthogonal projection approach \cite{miles2024vkd} to guide student network learning MAV motions. Unlike the original method \cite{miles2024vkd} (illustrated in Fig. \ref{fig:KDfigure}), this work proposes a stage-wise knowledge distillation strategy, which extends the orthogonal projection approach \cite{miles2024vkd} to the feature representations at each stage. Specifically, the feature maps from every stage of the teacher backbone are first standardized, following the approach in a work \cite{ying2021psigmoid}, after which the orthogonal projection of the corresponding student features is computed. The final distillation loss is then obtained based on these projection results.

To enhance the transfer of robust knowledge for MAV action recognition, we propose a stage-wise knowledge distillation framework that extends the orthogonal projection approach \cite{miles2024vkd} across multiple stages of the backbone networks. Utilizing ResNet18 as the teacher and MobileNetV4 as the student, our model distills intermediate features from stages C2 to C5, enabling the student to progressively learn hierarchical representations of MAV motions, such as dynamic trajectories and spatial maneuvers. For each stage $ l $ (where $ l $ ranges from C2 to C5), we align the student's features $ \mathbf{Z}_l^s \in \mathbb{R}^{b \times d_l^s} $ with the teacher's features $ \mathbf{Z}_l^t \in \mathbb{R}^{b \times d_l^t} $, where $ b $ is the batch size and $d_l^t$ is the dimension of the feature vector of the teacher network at stage $l$. If dimensions differ, the student features are first adapted via a learnable linear transformation to match a common dimension $ d_l $ in Eq. \ref{eq:1}.
\begin{equation}
\begin{split}
& \mathbf{F}_l^s = \mathbf{A}_l (\mathbf{Z}_l^s), \quad \\
& \mathbf{A}_l : \mathbb{R}^{d_l^s} \to \mathbb{R}^{d_l}.
\label{eq:1}
\end{split}
\end{equation}
Next, an orthogonal projection $P$ is applied to preserve the intra-batch similarity of student features in Eq. \ref{eq:2}, ensuring structural integrity crucial for distinguishing subtle MAV actions.
\begin{equation}
\begin{split}
& \mathbf{\tilde{F}}_l^s = \mathbf{F}_l^s \mathbf{P}_l, \quad \\
& \mathbf{P}_l = \phi(\mathbf{W}_l)_{:d_l}, \quad \\
& \phi(\mathbf{W}_l) = \exp(\mathbf{W}_l).
\label{eq:2}
\end{split}
\end{equation}
Here, $ \mathbf{W}_l $ is a skew-symmetric matrix derived from a learnable parameter, and the exponential is approximated efficiently (e.g., via Taylor series) to enforce $ \mathbf{P}_l \mathbf{P}_l^T = \mathbf{I} $, maintaining inner products in Eq. \ref{eq:3}.
\begin{equation}
\begin{split}
\langle \mathbf{\tilde{F}}_{l,i}^s,  \mathbf{\tilde{F}}_{l,j}^s \rangle = \langle \mathbf{F}_{l,i}^s, \mathbf{F}_{l,j}^s \rangle.
\label{eq:3}
\end{split}
\end{equation}

To improve convergence and robustness against input variations common in MAV video sequences, in Eq. \ref{eq:4}, the teacher features $\mathbf{\tilde{Z}}_l^t$ are standardized at each stage $l$.
\begin{equation}
\begin{split}
\mathbf{\tilde{Z}}_l^t = \frac{\mathbf{Z}_l^t - \mu(\mathbf{Z}_l^t)}{\sigma(\mathbf{Z}_l^t) + \epsilon},
\label{eq:4}
\end{split}
\end{equation}
where $\mu$ is mean operator and $\sigma$ is the variance operator. Besides, $\epsilon$ is a small positive constant. The stage-wise distillation loss is then computed as the L2 distance between the projected student features and standardized \cite{sun2024logit} teacher features, aggregated across all stages in Eq. \ref{eq:5}.
\begin{equation}
\begin{split}
\mathcal{L}_{\text{distill}} = \sum_{l=\text{C2}}^{\text{C5}} \left\| \mathbf{\tilde{F}}_l^s - \mathbf{\tilde{Z}}_l^t \right\|_2^2.
\label{eq:5}
\end{split}
\end{equation}
This hierarchical approach ensures the student captures multi-level MAV action cues, from low-level motion patterns in early stages to high-level semantic representations in later ones, without introducing excessive computational overhead.

\subsection{Similarity-Aware Attention Module}
To enhance the motion perception capability of our proposed method for MAV action videos, we introduce a similarity-aware attention (SAA) module to focus on the spatiotemporal information of MAV motion, thereby further improving the model’s discriminative ability for MAV actions in videos. The SAA module is a parameter-free attention mechanism that exploits the local self-similarity of spatiotemporal feature maps. Unlike conventional attention modules with extra learnable parameters, SAA adaptively enhances motion-relevant regions while suppressing background noise, making it well-suited for MAV video action recognition where lightweight and efficient models are required. 

Given a spatiotemporal feature map $X \in \mathbb{R}^{B \times C \times T \times H \times W}$ extracted from a video, SAA measures the distinctiveness of each feature map pixel $x_{t,i,j}$ by computing its deviation from the surrounding spatiotemporal neighborhood $\Omega_{t,i,j}$ in Eq. \ref{eq:s_att}.
\begin{equation}
\label{eq:s_att}
S_{t,i,j} = \frac{1}{N} \sum_{k \in \Omega_{t,i,j}} (x_{t,i,j} - x_k)^2 ,
\end{equation}
where $t$ denotes the temporal index, $(i,j)$ the spatial location, and $N$ the number of neighbors in the spatiotemporal neighborhood. Next, the similarity score is then normalized into an attention weight $w_{t,i,j}$ in Eq. \ref{eq:w_att}.
\begin{equation}
w_{t,i,j} = \frac{1}{1 + \exp\!\Big(-\tfrac{1}{4}\big(\tfrac{S_{t,i,j}}{\sigma^2+\epsilon}-1\big)\Big)} ,
\label{eq:w_att}
\end{equation}
where $\sigma^2$ is the estimated variance of similarity values and $\epsilon$ is a small constant for stability.  

Finally, the spatiotemporal attention map $W \in \mathbb{R}^{B \times 1 \times T \times H \times W}$ is applied element-wise to the original feature map $X' = W \odot X $. Through this mechanism, SAA dynamically emphasizes temporal motion cues (e.g., MAV trajectory and body orientation changes across frames) while suppressing static or irrelevant background information, thereby improving the accuracy of MAV video action recognition under complex environments.

\subsection{Hybrid Training Loss for Final Training}

\begin{figure}
\centering 
\includegraphics[width=0.49\textwidth]{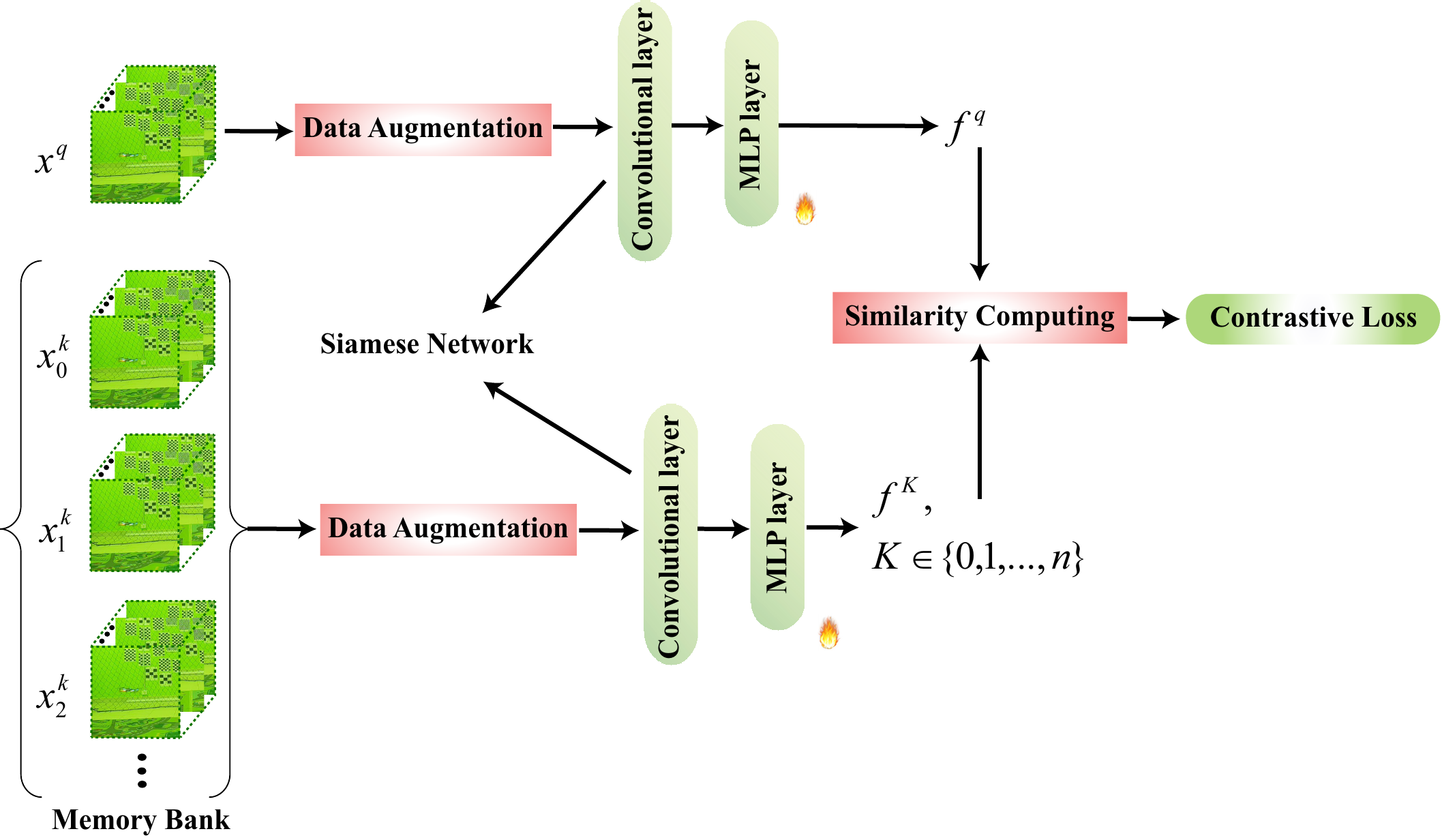}
\caption{Basic principles of contrastive learning of MAV action videos, where $x$ is MAV action sample and $f^q,f^K$ represent latent vector features. } \label{fig:CFfigure}
\end{figure}

For final network training, we design a set of loss functions, including distillation-related and contrastive learning objectives. Contrastive learning is particularly beneficial because labeled MAV video data is often limited and expensive to collect. With only limited labeled data available, contrastive learning provides an effective way to strengthen the network’s understanding of video MAV action recognition. In addition to contrastive learning, we further enhance knowledge transfer through distillation-based objectives. To enhance the performance of the lightweight MAV action recognition network, our objective is to improve action classification accuracy and align the classification distribution of the student network with that of the teacher network through multi-stage orthogonal knowledge distillation and soft-label knowledge distillation. To this end, the hybrid loss function of the network consists of four components: the cross-entropy loss for action classification, contrastive learning loss, the multi-stage orthogonal knowledge distillation loss, and the soft-label knowledge distillation loss.

As for contrastive learning (illustrated in Fig. \ref{fig:CFfigure}), it builds representations by encouraging similarity between temporally consistent views of the same action (positive pairs) while pushing apart unrelated actions (negative pairs). This mechanism allows the network to capture discriminative motion cues across frames in MAV videos. The process can be formulated using the InfoNCE loss \cite{he2020momentum} and it describes in Eq. \ref{eq:info}.  
\begin{equation}
\mathcal{L}_{\text{con}} = -\log \frac{\exp(\text{sim}(q, k^+)/\tau)}{\sum_{i=0}^N \exp(\text{sim}(q, k_i)/\tau)} ,
\label{eq:info}
\end{equation}
where $\text{sim}(\cdot)$ denotes feature similarity (e.g., dot product), $q$ is the query feature, $k^+$ is the positive 
sample from the same action sequence, $k_i$ are negative samples, and $\tau$ is a temperature parameter. By optimizing 
$\mathcal{L}_{\text{con}}$, the model learns temporal invariances that are crucial for robust MAV action recognition.  

To achieve our final goal of training a lightweight yet accurate model, we integrate contrastive learning with supervised classification and stage-wise orthogonal knowledge distillation into a unified training objective in Eq. \ref{eq:L_total}.  
\begin{equation}
\mathcal{L}_{\text{total}} = \alpha \mathcal{L}_{\text{CE}} + \beta \mathcal{L}_{\text{con}} + \gamma \mathcal{L}_{\text{distil}} + \delta \mathcal{L}_{\text{KD}},
\label{eq:L_total}
\end{equation}
where $\mathcal{L}_{\text{CE}}$ ensures correct classification, $\mathcal{L}_{\text{con}}$ facilitates the learning of robust MAV action representation, and $\mathcal{L}_{\text{distil}}$ aligns intermediate feature spaces via orthogonal projection from the teacher network. Besides, $\mathcal{L}_{\text{KD}}$ is traditional logit-based knowledge distillation. The coefficients $\alpha$, $\beta$, $\gamma$, and $\delta$ control the contribution of each term. This multi-objective design allows the student network to simultaneously learn discriminative features, benefit from the teacher’s knowledge, and remain lightweight, making it well-suited for MAV action recognition.  

\begin{figure}
\centering 
\includegraphics[width=0.49\textwidth]{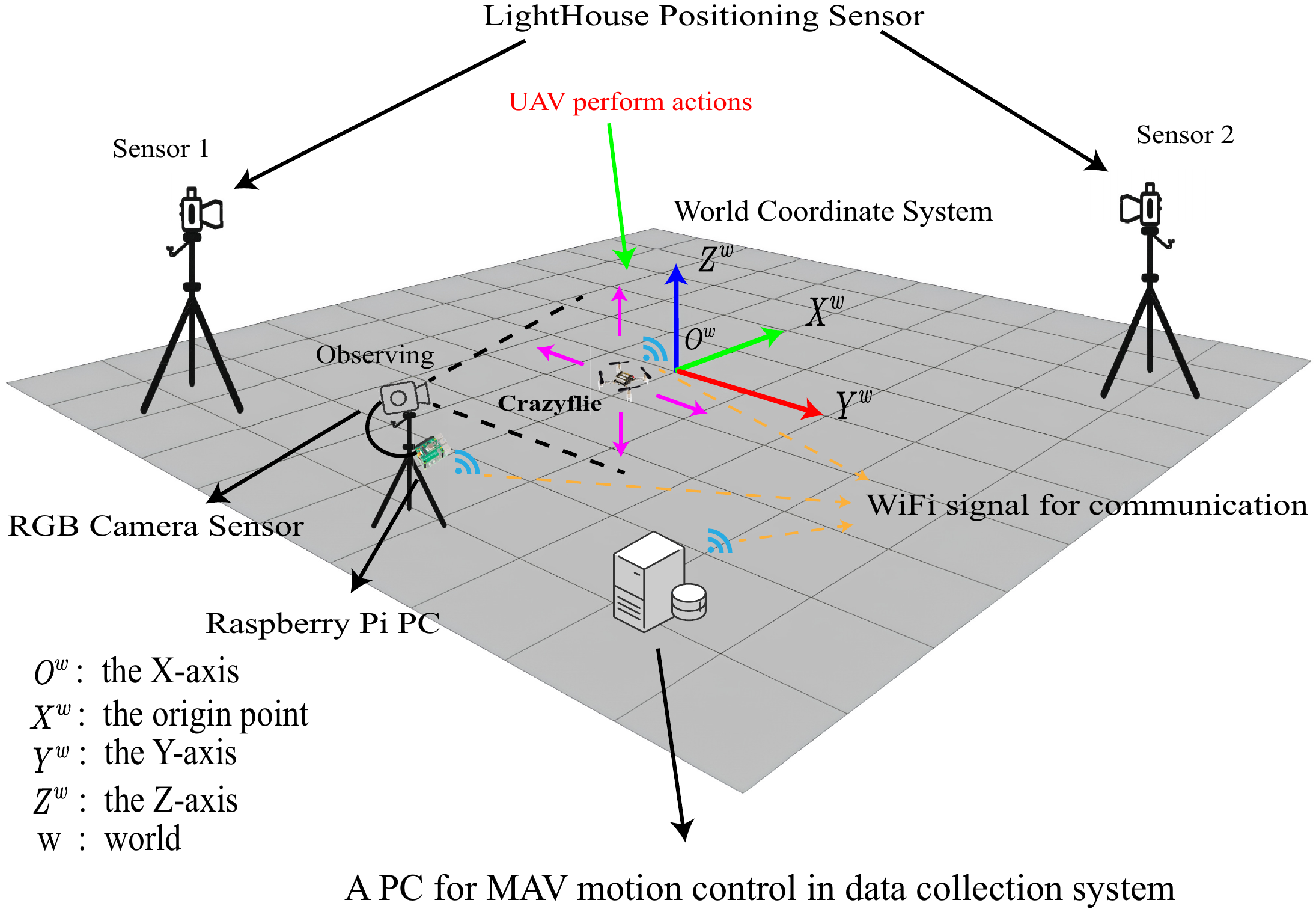}
\caption{Overview of collection scenarios for the MAV action recognition dataset.} 
\label{fig:dataCollection}
\end{figure}

\section{EXPERIMENT}
In this section, we conduct extensive comparative experiments to highlight the efficiency and other significant advantages of the proposed model. First, the experimental data, settings, and acquisition environment are described in Section A. Then, Section B presents a comparison between our model and other approaches in terms of MAV action recognition performance. The impact of the proposed SOKD module is evaluated in Section C. In Section D, we provide detailed ablation studies on the proposed modules. Finally, Sections E and F report the computational efficiency of the model as well as its power consumption and latency during single-sample inference. The dataset and codes used in this paper are available online\footnote{The dataset and source code used in this study will be made publicly available upon the publication of this paper.}.


\subsection{Dataset Introduction and Experimental Setting }
To evaluate the effectiveness of our model, we collected three MAV action datasets at different scales in a laboratory environment (illustrated in Fig. \ref{fig:dataCollection}). These datasets are designed to capture the motion of MAV at short, medium, and long distances, corresponding to flight distances of 1 m, 1.5 m, and 2 m from the RGB camera sensor. By analyzing actions across different scales, the performance of MAV action recognition algorithms can be assessed more intuitively. In our experiments, the MAV platform is Crazyflie 2.1, a lightweight and versatile micro aerial vehicle that supports precise motion control under the MAV link protocol. The data collection system is built on a Raspberry Pi 4.0 development kit, which provided sufficient computational resources for sensor integration and data processing. The RGB sensor had a resolution of 640 × 480 and ensured stable video capture. In addition, the Lighthouse tracking system is employed to record the real-time position of the MAV in the air. The position data are received by the Robot Operating System (ROS), which enabled communication between the tracking system and the Crazyflie MAV. Based on the Lighthouse data, ROS send control commands to the MAV, ensuring precise movements and synchronized action collection during the experiments. This integrated setup allowed us to capture high-quality datasets under controlled conditions, providing a solid foundation for evaluating the proposed action recognition model.

For training the proposed MAV action recognition model, we conducted experiments on a high-performance computing platform equipped with an NVIDIA RTX 4090 GPU, ensuring efficient training and inference. The software environment is based on Ubuntu 20.04, and the neural network models are developed and trained using the PyTorch deep learning framework. During preprocessing, MAV flight videos are trimmed and segmented using FFmpeg. Each MAV action is divided into independent clips, and the processed data are stored in HDF5 format to enable efficient access during training. For training configuration, the learning rate is set to 0.0001, and the number of training epochs is set to 25. The training dataset is split into training and testing sets with a ratio of 3:1. To assess the robustness and reliability of the model, we repeated some experiments ten times for each configuration and reported both the mean performance and the standard deviation to evaluate consistency.

\subsection{The Performance Comparison among Other State-of-the-art Model }

\begin{figure}[t]
\centering 
\includegraphics[width=0.45\textwidth]{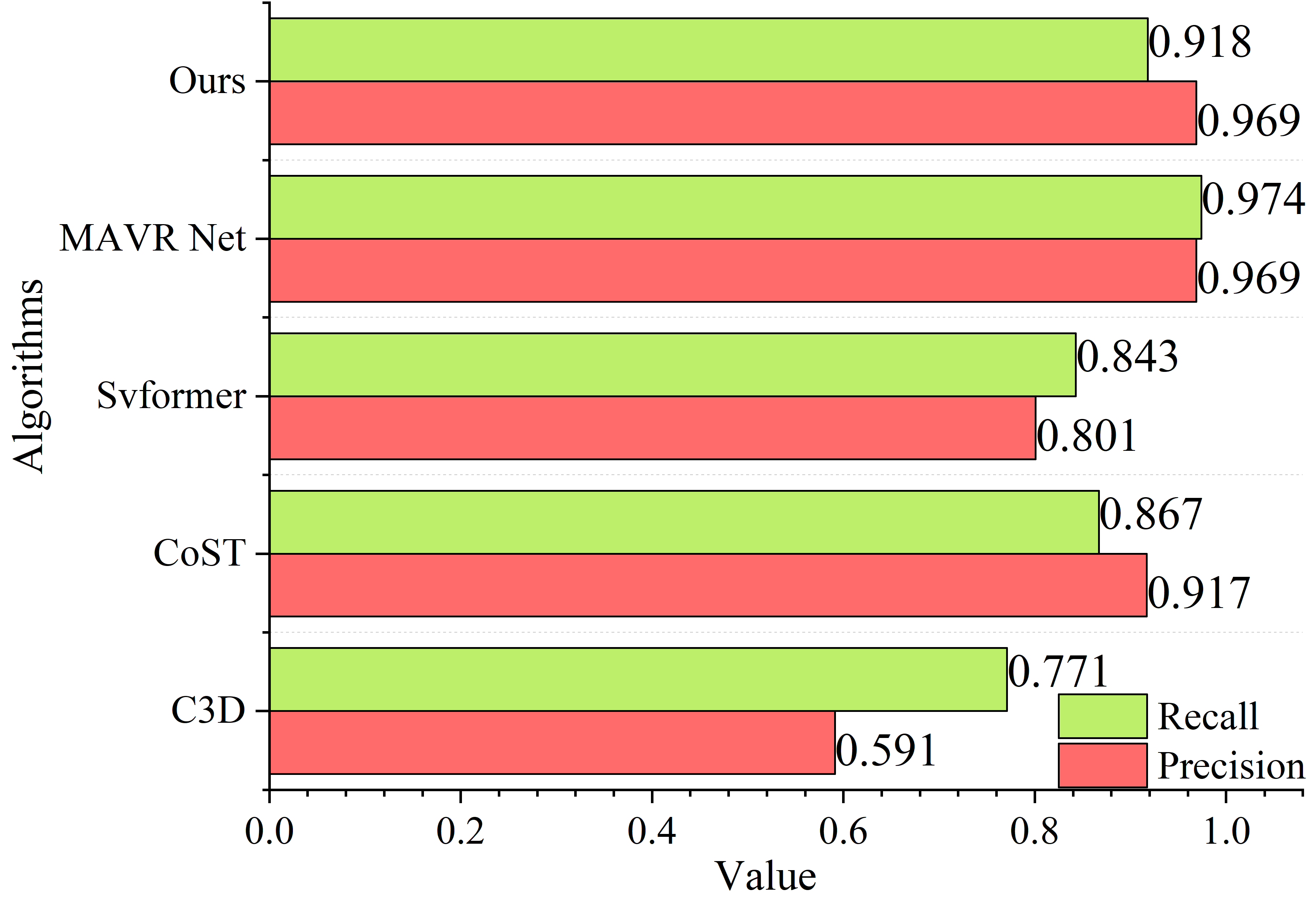}
\caption{The comparison of precision and recall in medium MAV action recognition dataset.} \label{fig:pre_recall}
\end{figure}

To validate the advanced nature of our proposed model in MAV action recognition task, we conducted experiments on three datasets (Short, Medium, and Long) with other state-of-the-art models (C3D\cite{tran2015learning}, CoST\cite{li2019collaborative}, Svformer \cite{xing2023svformer}, MAVR-Net \cite{zhang2025mavrnetrobustmultiviewlearning}). The experimental results are shown in Table \ref{tab:compar}. To ensure the reproducibility of the results, each algorithm is tested ten times, and the standard deviation is recorded.   

\begin{table}[h]
    \centering
    \caption{RESULTS OF MAV ACTION RECOGNITION METHODS IN COMPARISON.}
    \label{tab:compar}
    \begin{tabular}{ccccccc}
    \hline
    Options & Short & Medium & Long \\ \hline
     C3D &$70.4\%\pm 7.8\%$&$64.5\%\pm 8.4\%$   &$62.1\%\pm 16.4\%$  \\ 
    CoST &$87.5\%\pm 9.5\%$&$87.1\%\pm 11.1\%$ &$84.2\%\pm 13.5\%$ \\
    Svformer   &$89.1\%\pm 10.4\%$&$82.2\%\pm 15.1\%$ &$70.4\%\pm 19.7\%$ \\
   MAVR-Net  &$97.8\%\pm 2.8\%$&$96.5\%\pm 3.9\%$ &$92.8\%\pm 4.7\%$ \\
   ours  &$94.8\%\pm 4.7\%$&$92.84\%\pm 5.6\%$ &$88.71\%\pm 7.4\%$ \\
    \hline
    \end{tabular}
\end{table}

From Table \ref{tab:compar}, we can observe that as the distance between the MAV and the RGB sensor increases (from Short datasets to Long datasets), all models experience a decline in MAV action recognition accuracy. This is due to the smaller imaging scale of the MAV at greater distances, making action recognition for small targets less robust. However, our proposed lightweight model demonstrates highly competitive performance on MAV recognition tasks at all three scales. This result indicates that our lightweight model is well capable of recognizing MAV motion patterns at different scales. Besides, we employ precision and recall as evaluation metrics for MAV action recognition, as these metrics better capture the effects of class imbalance and the model’s ability to correctly identify diverse actions in complex aerial environments. Precision assesses prediction accuracy, while recall evaluates the completeness of identification. The specific experimental results are shown in Fig. \ref{fig:pre_recall}. The bar chart compares the performance of five algorithms on the medium MAV action dataset in a evaluation: Svformer with precision 0.801 and recall 0.843, CoST with 0.917 and 0.867, C3D with 0.591 and 0.771, MAVR-Net with 0.969 and 0.974. As a lightweight model, the proposed method achieves precision and recall value of 0.969 and 0.918, respectively, delivering highly competitive MAV recognition performance.

\begin{figure}
\centering 
\includegraphics[width=0.43\textwidth]{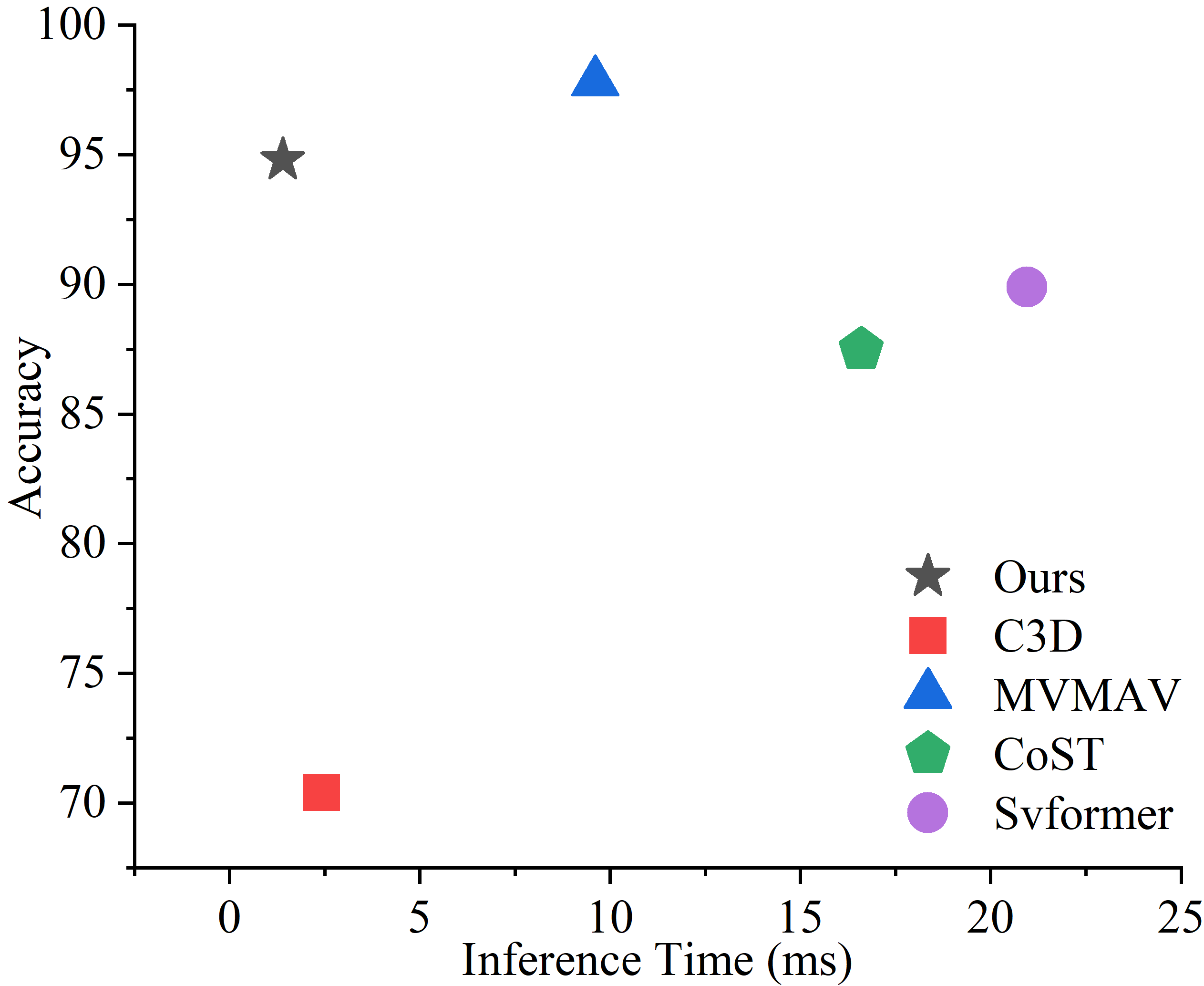}
\caption{Comparative analysis of accuracy and inference time across different MAV action recognition models.} \label{fig:inf_acc}
\end{figure}

\begin{figure*}[h]
\centering 
\subfigure[Short dataset.]{%
\resizebox*{5.5cm}{!}{\includegraphics{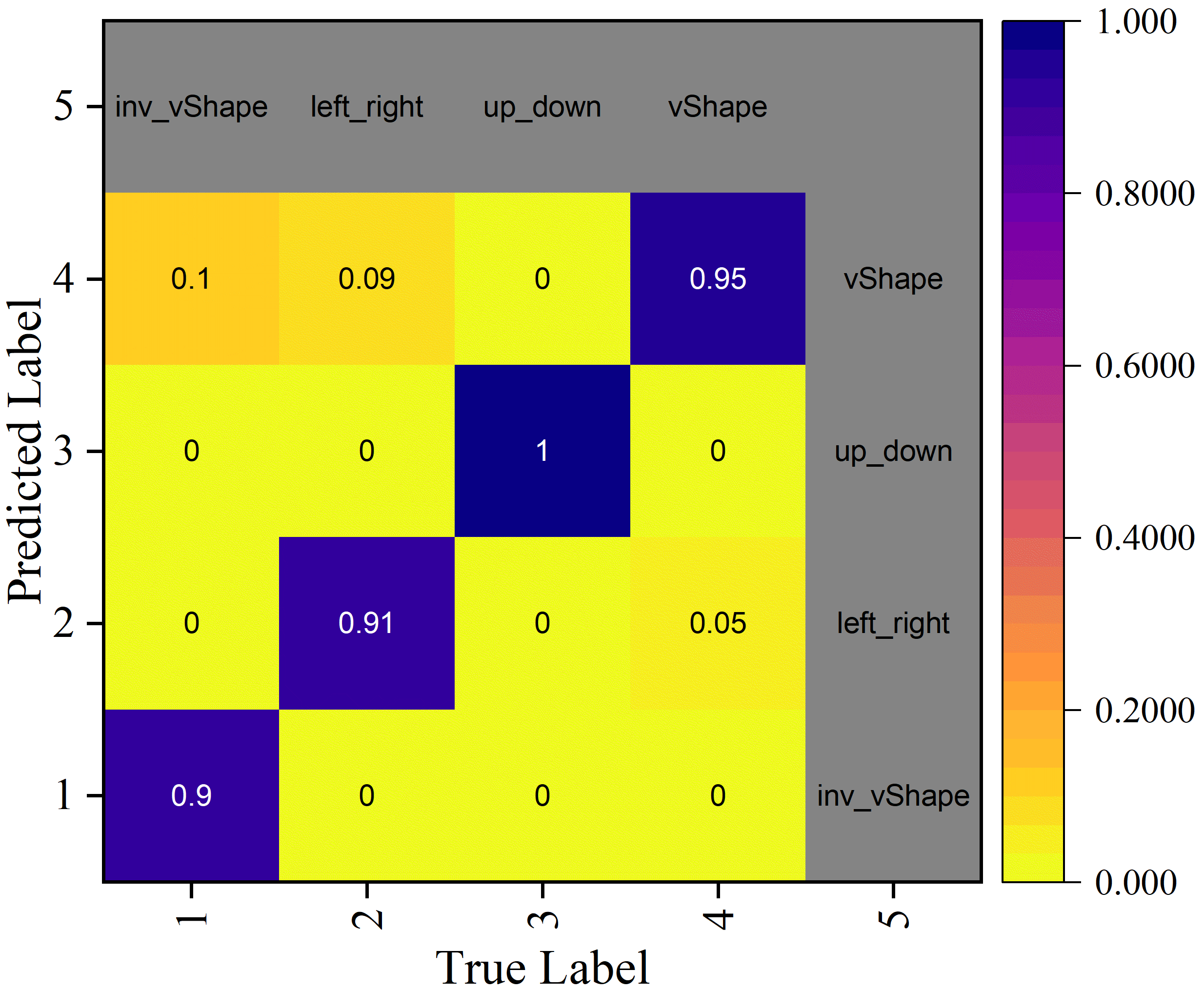}}}
            \hspace{0.05cm}
\subfigure[Medium dataset.]{%
\resizebox*{5.5cm}{!}{\includegraphics{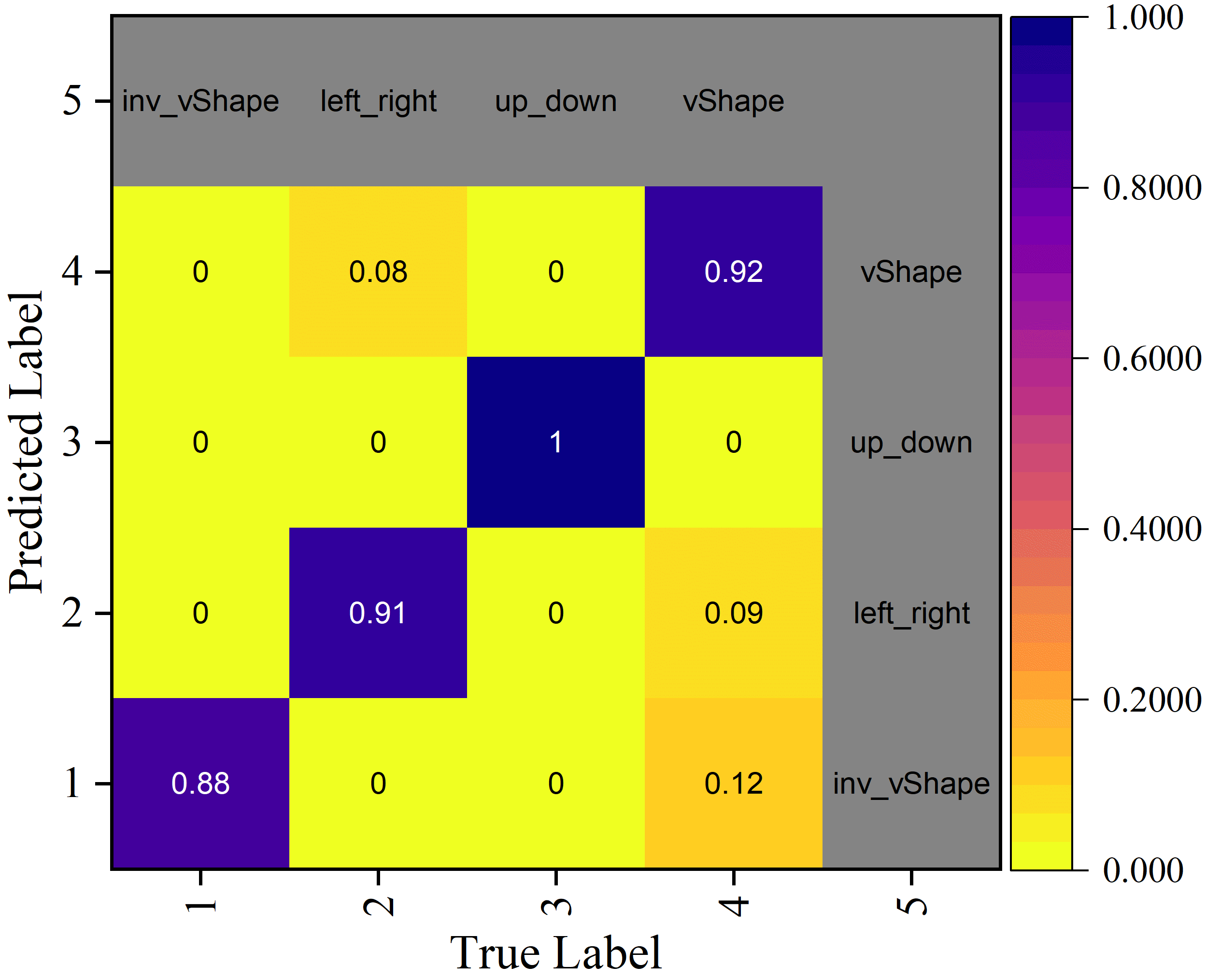}}}
           \hspace{0.05cm}
\subfigure[Long dataset.]{%
\resizebox*{5.5cm}{!}{\includegraphics{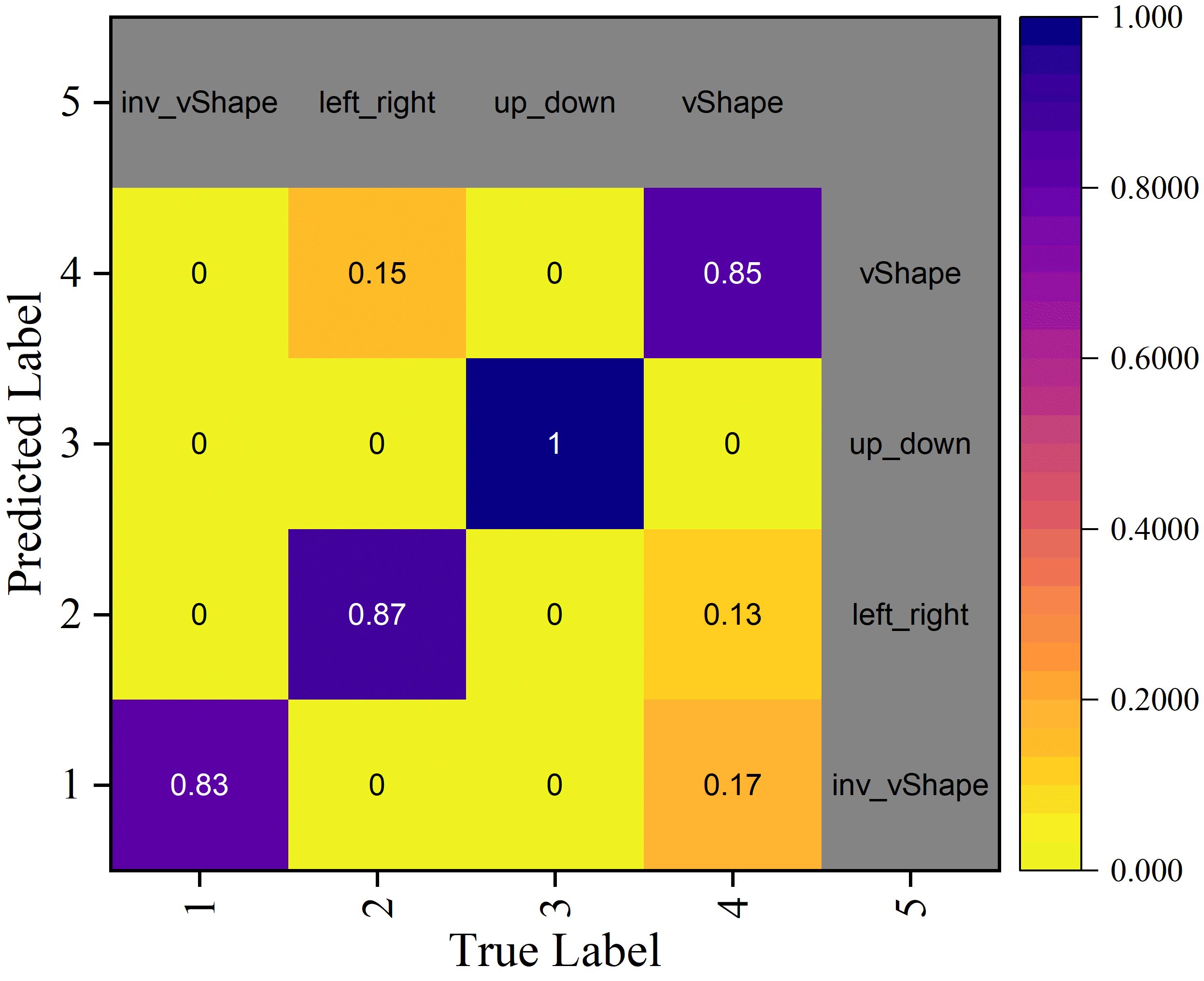}}}
\caption{Confusion matrices for Short (a), Medium (b), and Long (c) MAV action recognition datasets.} \label{fig:confusion}
\end{figure*}

To demonstrate the recognition efficiency of our proposed model, we present a comparison of inference speed and recognition accuracy with other models in Short MAV action recognition dataset, as shown in Fig. \ref{fig:inf_acc}. The horizontal axis represents the average inference time per sample on an NVIDIA 4090 GPU. Clearly, both C3D and our model achieve very low inference times, below 5 ms per sample. However, our model maintains highly competitive recognition accuracy with minimal latency. This highlights its excellent balance between efficiency and performance. Besides, we also provide the confusion matrix of the proposed model on the MAV action recognition dataset at three different scales (as shown in Fig. \ref{fig:confusion}). It shows that the proposed model has good recognition results for various MAV action classes.

\begin{table}[h]
    \centering
    \caption{The influence of the teacher in the SOKD method for MAV action recognition.}
    \label{tab:kd_teacher}
    \begin{tabular}{ccccccc}
    \hline
    Teacher & Short & Medium & Long \\ \hline
    Resnet18 &$94.8\%\pm 4.7\%$&$92.84\%\pm 5.6\%$   &$88.71\%\pm 7.4\%$  \\ 
    ConvNeXt &$95.05\%\pm 3.5\%$&$92.94\%\pm 4.1\%$ &$88.92\%\pm 6.5\%$ \\
    ViT   &$93.91\%\pm 2.8\%$&$92.2\%\pm 4.7\%$ &$87.64\%\pm 6.7\%$ \\
    \hline
    \end{tabular}
\end{table}

\subsection{The Impact of Stage-wise Orthogonal Knowledge Distillation in different stages and different teacher  }
\begin{figure}[t]
\centering 
\includegraphics[width=0.38\textwidth]{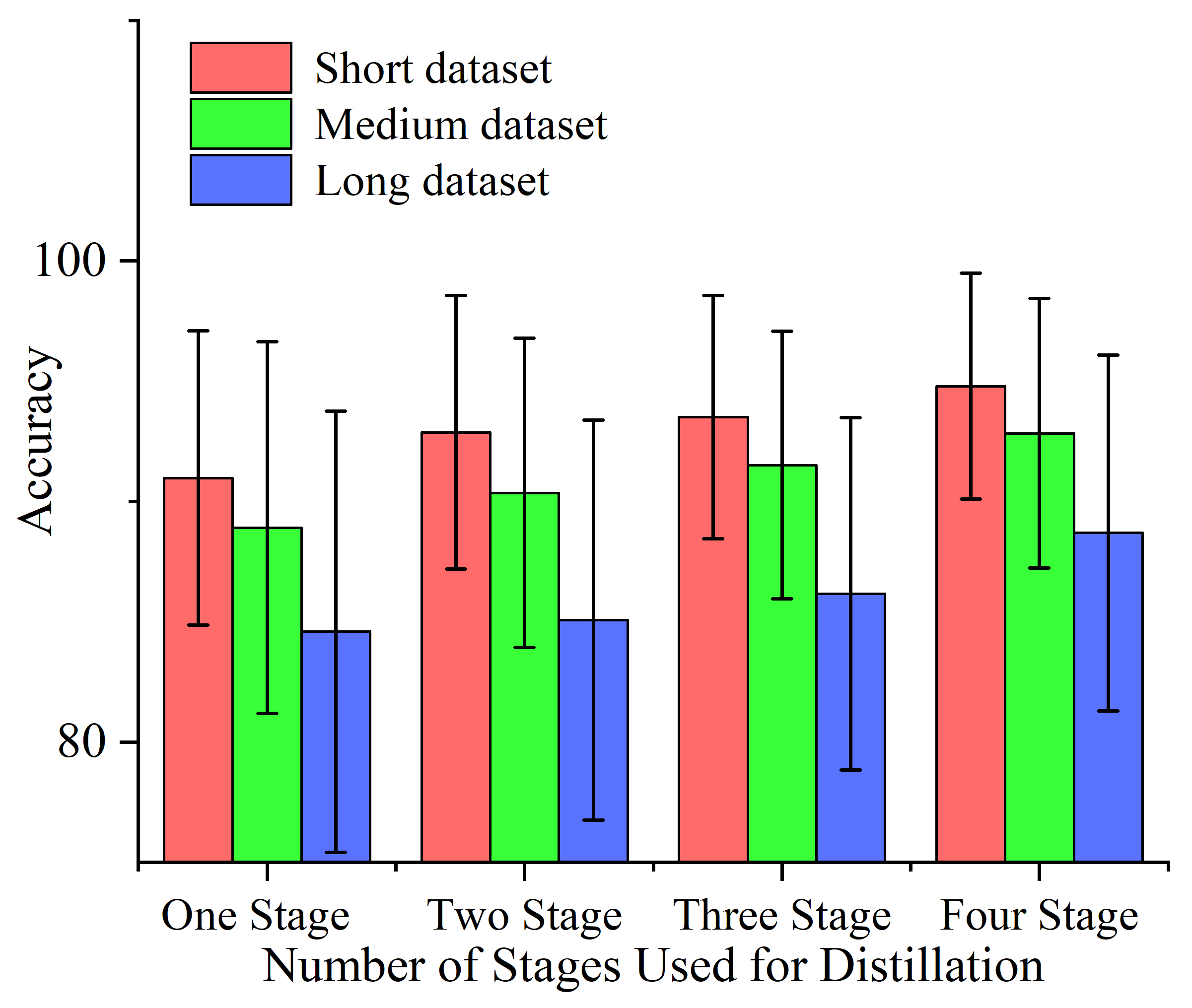}
\caption{The impact of the number of stages on MAV action recognition.} \label{fig:stage_dis}
\end{figure}

To investigate the impact of multi-stage distillation on the performance of MAV action recognition, we conducted experiments on the short, medium, and long MAV action recognition datasets. Specifically, we distilled the orthogonal projections of the first $N$ feature stages from the ResNet18 backbone into the MobileNetV4 backbone. Initially, we used only the C2 stage features for distillation (One Stage). Then, we included both C2 and C3 stages (Two Stage), and gradually increased the number of stages up to four (C2, C3, C4, and C5). Each experiment is repeated ten times, and the average accuracy and standard deviation of the student network are reported, as shown in Fig.~\ref{fig:stage_dis}. The results demonstrate that incorporating more stages in the distillation process progressively improves the recognition accuracy of the student model for MAV actions. Furthermore, by introducing additional features in SOKD, the training outcomes become more stable, as reflected by the reduced standard deviation of MAV action recognition results.

\begin{table*}[t]
    \centering
    \caption{Ablation experiments on the proposed model.}
    \label{tab:ablation}
    \begin{tabular}{ccccccc}
    \hline
    Options & SOKD  & Contrastive loss & Parameter-free attention                      & Short              & Medium             & Long \\ \hline
    SOKD  & \checkmark &- & -                                 &$91.36\%\pm 6.4\%$  &$89.3\%\pm 7.8\%$   &$86.3\%\pm 8.7\%$  \\ 
    SOKD+CL  & \checkmark&\checkmark & -                     &$92.8\%\pm 5.1\%$   &$91.15\%\pm 6.6\%$ &$88.8\%\pm 8.4\%$ \\
    SOKD+CL+Attention  &\checkmark & \checkmark&\checkmark   &$94.8\%\pm 4.7\%$   &$92.84\%\pm 5.6\%$   &$88.71\%\pm 7.4\%$ \\
    \hline
    \end{tabular}
\end{table*}

To investigate the influence of different teacher models on the SOKD framework and their impact on the student network’s MAV action recognition performance, we conducted comparative experiments under various teacher settings based on the fundamental principles of knowledge distillation (as illustrated in Fig. \ref{fig:KDfigure}). Specifically, three types of candidate teacher networks are selected, ResNet18 \cite{he2016deep}, ConvNeXt \cite{liu2022convnet}, and Vision Transformer (ViT) \cite{yuan2021tokens}, to systematically evaluate their effect on the distillation process. Each experiment is repeated ten times, and the average recognition accuracy as well as the standard deviation of the student model are reported.

As shown in Table \ref{tab:kd_teacher}, when ConvNeXt is employed as the teacher model, the student network achieved the highest classification performance in MAV action recognition, with the most stable results. In contrast, when the more powerful Vision Transformer is used as the teacher, the classification accuracy of the student model slightly decreased, although the stability of the results improved. The decrease in accuracy may be attributed to the architectural differences between the teacher and the student networks, which lead to divergent perspectives in feature extraction for MAV actions and may limit the effectiveness of knowledge distillation. This observation is further supported by the findings in a work \cite{hao2023one}. In summary, the proposed SOKD method demonstrates robust effectiveness in transferring MAV action knowledge across different teacher architectures.

\subsection{Ablation Experiments in The Proposed Model for MAV Action Recognition}

To assess the contribution of each module to MAV recognition tasks, we conducted ablation studies on the proposed model, as detailed in Table \ref{tab:ablation}. We compared the performance of the lightweight Mobile MAV model across three configurations: the SOKD baseline model, the SOKD + Contrastive Loss (CL) model, and the final SoKD + CL + Attention model. The results clearly demonstrate the incremental benefits of each component. 

The SOKD baseline model, relying solely on Stage-wise Orthogonal Knowledge Distillation, achieved respectable performance with scores of $91.36\%\pm 6.4\%$ (Short), $89.3\%\pm 7.8\%$ (Medium), and $86.3\%\pm 8.7\%$ (Long). Incorporating CL in the SOKD + CL model significantly improved recognition accuracy and stability, yielding scores of $92.8\%\pm 5.1\%$ (Short), $91.15\%\pm 6.6\%$ (Medium, and $88.8\%\pm 8.4\%$ (Long). This represents an average improvement of approximately $1.5\%$ to $2\%$ across three datasets, highlighting the effectiveness of CL in enhancing MAV action recognition. The final SOKD + CL + Attention model, which integrates a parameter-free attention module, delivered the best overall performance, with scores of $94.8\%\pm 4.7\%$ (Short), $92.84\%\pm 5.6\%$ (Medium), and $88.71\%\pm 7.4\%$ (Long). This marks a further improvement of about 2\% to 3\% over the SOKD + CL model, particularly noticeable in the Short and Medium datasets. Interestingly, after introducing the attention mechanism, a performance decline was observed on the Long dataset. This is primarily because the MAVs in the Long dataset were captured at a greater distance from the sensor, resulting in smaller image scales. Under such conditions, even with the attention mechanism, the model struggles to capture subtle MAV motion variations. Overall, the results demonstrate that the lightweight model achieves satisfactory MAV action recognition performance using SOKD alone. By incorporating CL, the MAV action recognition capability is further enhanced, with improved stability in recognition results. In the final model, the introduction of a parameter-free attention module leads to the best MAV action recognition results.    

\begin{figure}[h]
\centering 
\subfigure[Accuracy: SOKD vs No SOKD.]{%
\resizebox*{4.2cm}{!}{\includegraphics{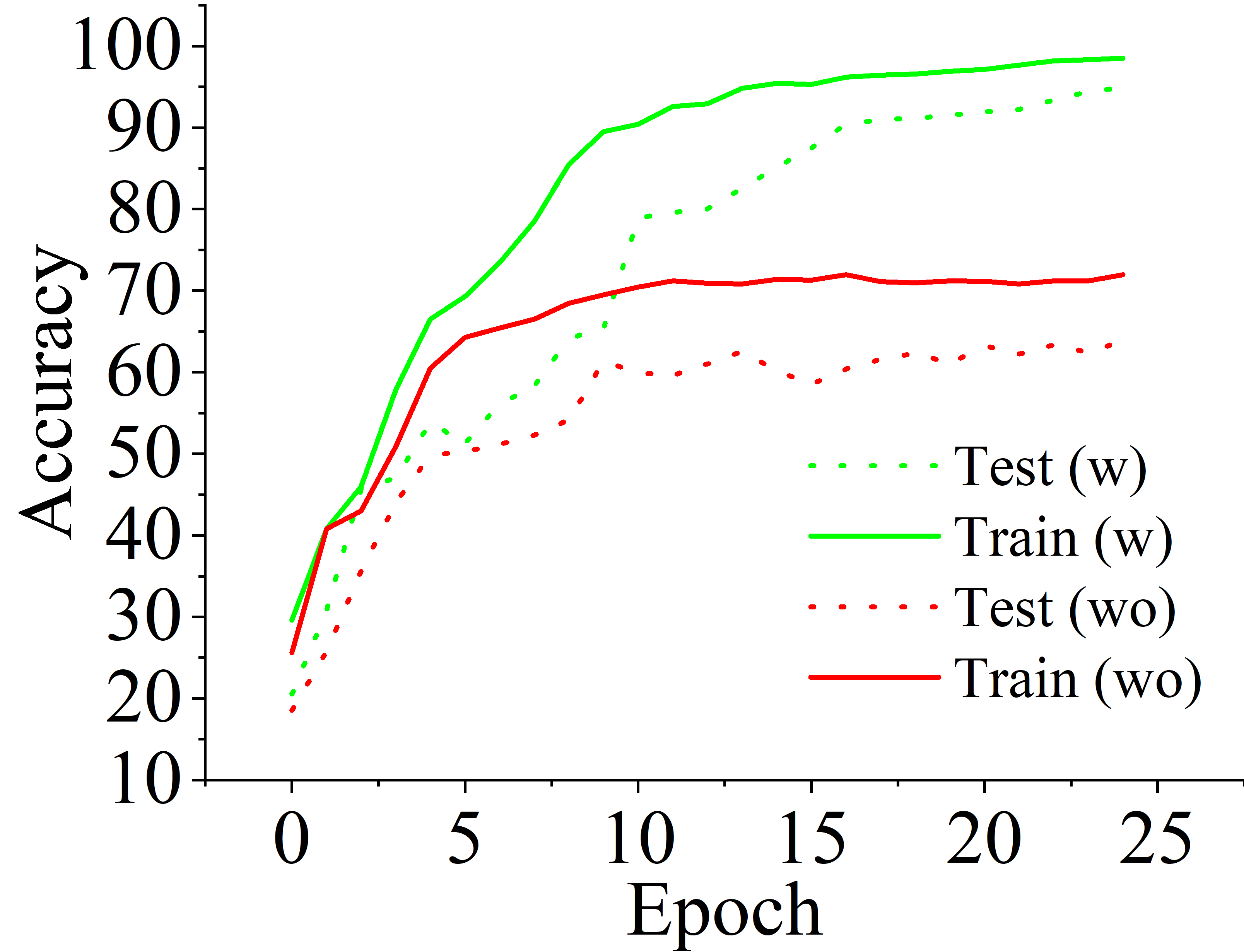}}}
            \hspace{0.05cm}
\subfigure[Loss: SOKD vs No SOKD.]{%
\resizebox*{4.2cm}{!}{\includegraphics{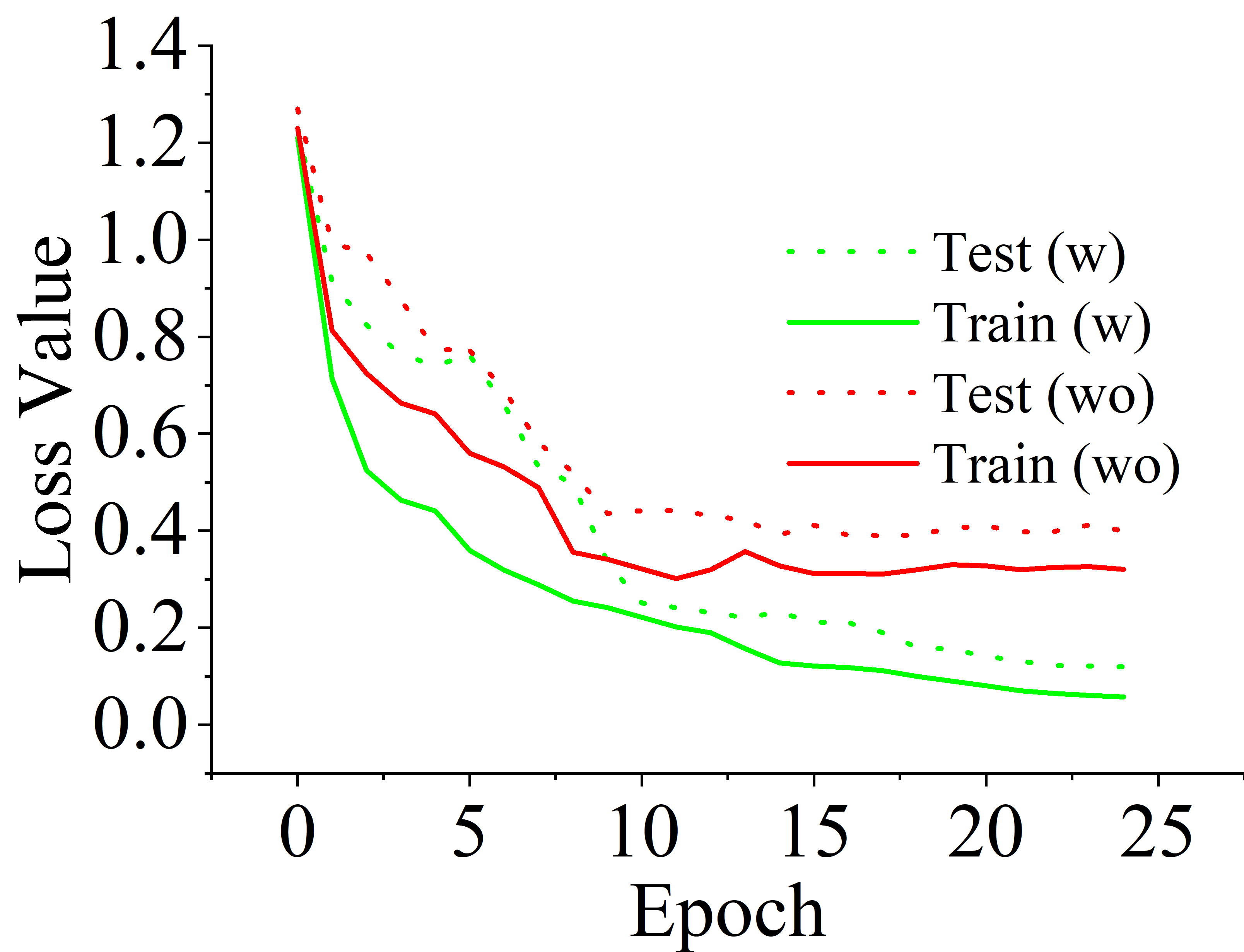}}}
\caption{The figure shows (a) Accuracy: SOKD vs No SOKD and (b) Loss: SOKD vs No SOKD, comparing model performance with and without Stage-wise Orthogonal Knowledge Distillation (SOKD) across epochs.} \label{fig:sokd}
\end{figure}


To validate the effectiveness of the Stage-wise Orthogonal Knowledge Distillation (SOKD) framework, its ablation study is presented in Fig. \ref{fig:sokd}. Specifically, we compare two settings: with SOKD (w) and without SOKD (wo), under the Short MAV action dataset, in terms of convergence behaviors and learning speed on both the training and testing sets. As illustrated in Fig. \ref{fig:sokd}, SOKD plays a crucial role in training lightweight models. When SOKD is applied, both accuracy and loss on the training and testing sets converge to more desirable states. In contrast, without SOKD, i.e., when the student network is trained independently, its accuracy saturates at around $60\%$ and no longer improves. A similar phenomenon can be observed in the loss curves. The primary reason is that, without the knowledge guidance from the teacher network, the student model struggles to effectively capture discriminative MAV action information solely by its own capacity. This observation is consistent with the conclusions reported in early studies on knowledge distillation \cite{gou2021knowledge}.

\subsection{Computational Efficiency Compared with Other State-of-the-art MAV Action Recognition Model}
Conducting a comprehensive comparison of computational efficiency is essential to validate the effectiveness of our proposed MAV action recognition model against existing state-of-the-art approaches. This evaluation is critical for identifying potential improvements in resource-constrained environments. The study assesses key computational metrics, including parameter count (M), floating-point operations (FLOPs in G), multiply-add operations (MACs in G), and model storage size (MB).

Table \ref{tab:Complexity} presents the comparative analysis, highlighting that our model achieves a parameter count of 11.34 M, 59.20 G FLOPs, 29.60 G MACs, and a storage size of 43.99 MB. In contrast, the leading models (C3D, Svformer, and MAVR-Net) exhibit higher values, with C3D at 78.01 M parameters and 297.60 MB size, Svformer at 14.59 M and 55.89 MB, and MAVR-Net at 79.90 M and 305.05 MB respectively. These results underscore the efficiency of our approach. Besides, it also demonstrates that our model can maintain high recognition accuracy while exhibiting low computational complexity and storage requirements. This efficiency paves the way for deploying lightweight, high-performance models in future MAV applications, addressing the critical need for resource optimization in real-world scenarios.

\begin{table}[h]
    \centering
    \caption{Computational Complexity Comparison for MAV Action Recognition Algorithms.}
    \label{tab:Complexity}
    \begin{tabular}{ccccccc}
    \hline
    Algorihtms & Parameters (M) & Flops (G) & MAC (G) & size (MB)\\ \hline
    C3D   &78.01& \textbf{38.66} & \textbf{19.33} & 297.60 \\ 
    CoST  &21.09&186.61 &93.07 &107.43\\
    Svformer   & 14.59& 209.57 &104.65 & 55.89 \\ 
    MAVR-Net  &79.90&128.65 &64.32 &305.05\\
     ours   & \textbf{11.34}& 59.20 &29.60& \textbf{43.99} \\ 
    \hline
    \end{tabular}
\end{table}

\subsection{Energy Cost and Inference Time Analysis on Single MAV Action Video Sample}

To gain a more intuitive understanding of the advantages of our proposed model in terms of energy consumption and inference efficiency, we conducted real-world hardware testing. We evaluated different models by measuring their inference latency across various computing environments, including CPU (Intel Xeon(R) Gold 6148 CPU 2.40GHz) and GPU (NVIDIA RTX 4090), and performed a comparative analysis. Additionally, we provided the estimated power consumption of different models. The energy consumption values are estimated based on the MAC operations of each model, using a conversion factor of 4.6 pJ per 1 GB on 45 nm CMOS hardware \cite{rathi2021diet}.

Based on the experimental results presented in Table \ref{tab:energy}, a detailed comparison of runtime and energy efficiency across various algorithms highlights the superior performance of our proposed model. The table provides key metrics including CPU time (s), GPU time (ms), decoding time (s), and energy consumption (pJ) for processing inputs of different resolutions. Specifically, the CoST algorithm, with an input of 224 x 224, exhibits a CPU time of 0.69 s, GPU time of 16.6 ms, decoding time of 1.44 s, and an energy consumption of 428.12 pJ. In contrast, the Svformer algorithm, processing 224 x 224 inputs, show a CPU time of 0.56 s, GPU time of 20.95 ms, decoding time of 1.78 s, and energy consumption of 481.39 pJ. Our model, designed for 128 × 128 inputs, achieves the lowest CPU time (0.113 s), GPU time (1.39 ms), and decoding time (8.84 s), while maintaining relatively low energy consumption (136.16 pJ) and improved inference efficiency compared with other models.  

The experimental results validate the effectiveness of our proposed model. By optimizing for 128 x 128 resolution inputs, our model not only reduces inference latency (evidenced by the minimal CPU and GPU times) but also achieves a substantial decrease in energy consumption. This is particularly evident when compared to the CoST and Svformer models, which consume 428.12 pJ and 481.39 pJ respectively, while our model uses only 136.16 pJ. It is worth noting that although C3D achieves the lowest energy consumption (88.91 pJ), its MAV action recognition accuracy is considerably lower than that of the proposed model. These findings underscore the model's efficiency in real-world hardware testing across diverse computing environments. The comprehensive evaluation confirms that our proposed model offers significant advantages in terms of energy efficiency and inference speed. Its low power consumption and rapid processing capabilities make it a highly effective solution for MAV action recognition tasks, positioning it as a superior choice in this domain. 

\begin{table}[h]
    \centering
    \caption{Runtime and Energy Efficiency Analysis of MAV Action Recognition Algorithms.}
    \label{tab:energy}
    \begin{tabular}{ccccccc}
    \hline
    Algorithms &Input & CPU  & GPU & Decoding  & Energy \\ 
    & (pixels) & (s) & (ms) & (/s) & (pJ) \\ \hline
    C3D   &112x112 & 0.122   & 2.41    &8.183  & \textbf{88.91}  \\ 
    CoST &224x224 & 0.69    & 16.6    &1.44  & 428.12 \\
    Svformer &224x224 & 0.56    & 20.95   &1.78  & 481.39  \\ 
    MAVR-Net &128x128 & 0.32    & 9.61    &3.11  & 295.87  \\
    ours  &128x128 & \textbf{0.113}   & \textbf{1.39}    & \textbf{8.84}  & 136.16   \\ 
    \hline
    \end{tabular}
\end{table}

\section{Conclusion}
In this paper, we propose a lightweight model for Micro Aerial Vehicle (MAV) action recognition. The proposed model accurately identifies four fundamental MAV motion patterns while maintaining high inference speed, low parameter complexity, and reduced power consumption, making it well-suited for deployment on embedded and mobile devices. Specifically, we employ a Stage-wise Orthogonal Knowledge Distillation (SOKD) framework to facilitate effective knowledge transfer between the teacher and student networks, enabling the student network to learn discriminative MAV action representations. In addition, a parameter-free attention mechanism is integrated to enhance MAV action capability without introducing extra training parameters, thereby keeping the model compact and memory-efficient. Finally, a hybrid training loss is adopted to ensure stable optimization during MAV action model training. Extensive experiments demonstrate that he proposed MAV action recognition model achieves faster inference (0.113 s per sample on CPU and 1.39 ms on GPU), while maintaining high accuracy with average recognition rates of 94.8\%, 92.84\%, and 88.7\% on the Short, Medium, and Long datasets, respectively. These results indicate that our method effectively balances accuracy and efficiency. In future work, we plan to further optimize the proposed model and extend it to recognize a broader range of MAV actions.


%


\section*{Acknowledgment}
This work was supported in part by [details omitted for double-anonymous review]. The authors will include complete acknowledgment information in the final version of the manuscript.

\ifCLASSOPTIONcaptionsoff
  \newpage
\fi



\bibliographystyle{IEEEtran}
\bibliography{IEEEexample}
\end{document}